# A Survey of Machine Learning Algorithms for Detecting Malware in IoT Firmware

Erik Larsen, Korey MacVittie, John Lilly

PeopleTec, Inc., 4901 Corporate Drive NW, Huntsville, AL 35805
erik.larsen@peopletec.com

## ABSTRACT

*This work explores the use of machine learning techniques on an Internet-of-Things firmware dataset to detect malicious attempts to infect edge devices or subsequently corrupt an entire network. Firmware updates are uncommon in IoT devices; hence, they abound with vulnerabilities. Attacks against such devices can go unnoticed, and users can become a weak point in security. Malware can cause DDoS attacks and even spy on sensitive areas like peoples' homes. To help mitigate this threat, this paper employs a number of machine learning algorithms to classify IoT firmware and the best performing models are reported. In a general comparison, the top three algorithms are Gradient Boosting, Logistic Regression, and Random Forest classifiers. Deep learning approaches including Convolutional and Fully Connected Neural Networks with both experimental and proven successful architectures are also explored.*

## 1. INTRODUCTION

The Internet-of-Things (IoT) is an evolution of networking and data synergized with connected edge devices. It promises a leap forward to integrating the connectivity of the internet with data that informs decisions, even down to mundane tasks like automatically ordering milk from a grocery store. This new realm of interconnected "smart" things is poised to make its way into every imaginable device, from home appliances to textiles. The introduction of myriad computing platforms all interconnected presents bad actors with opportunities to hack into peoples' belongings and subsequently into their lives. It has been estimated that by 2020 most people had, on average, four IoT devices [1]. Vu et al [2] find that 323,000 new malware files detected every day. Identifying malware will be essential in securing these networks as the technology matures.

Herein we build on the work of Ebbers [3] and Marais et al. [4] in exploring the syndication of machine learning and IoT firmware through the lens of cybersecurity. Results of these works expound upon the suitability for certain machine learning (ML) models to detect weaknesses and defend against attacks. Ebbers finds that the three main contributors to IoT firmware vulnerabilities are manufacturer, device type, and location (specifically country). Manufacturers have a tendency to choose rapid development over security; hence, devices tend to be riddled with vulnerabilities. Combine with this the erratic nature of firmware update releases and the reluctance of users to attempt installation – for lack of knowledge or technical ability – and it becomes common that devices are rarely updated or patched.

One previous approach [5] transforms 25 classes of malware into grayscale images. We take a similar tactic with IoT firmware as images but introduce only four distinct classes. Many different ML models have been used for similar tasks – e.g., the HIT method uses color image transformations – throughout the literature with remarkable success [2, 6, 7].

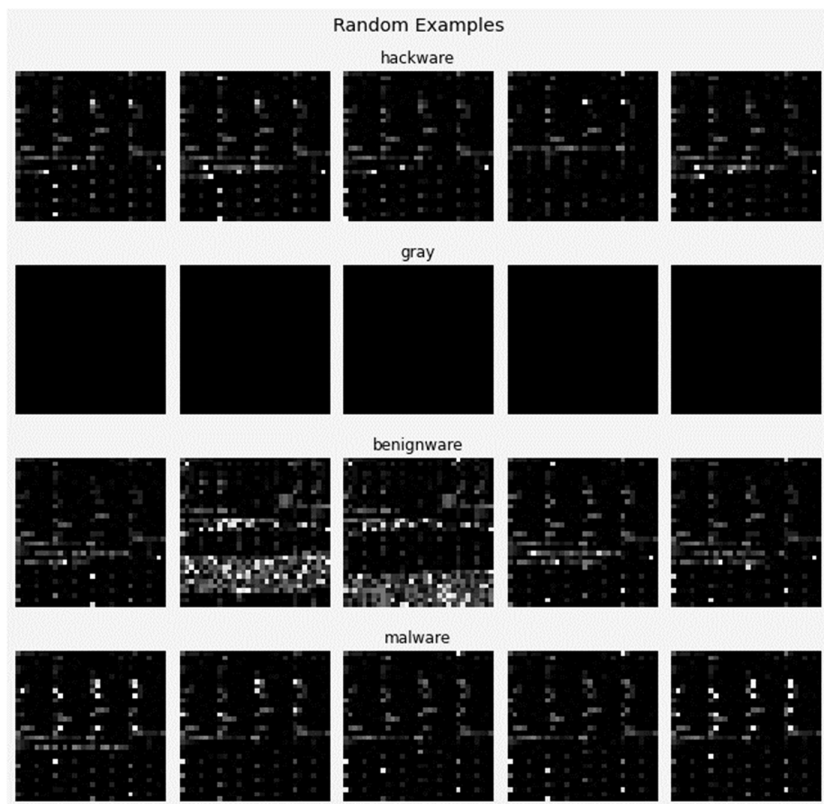

Figure 1.  IoT firmware malware as images includes "gray" for reference

Special attention is paid to minimizing false negatives due to the ramifications of misclassification, although packed or encrypted benign executables are often misclassified as malicious [3].

Results of cursory and exploratory data analyses (EDA) as well as initial trials with ML algorithms are presented. There is a great deal of research detailing methods for classifying the famous MNIST hand-written digit set using a plethora of machine learning techniques. This work hopes to leverage that vast knowledgebase in the realm of classifying images with the goal of "weeding out malware using image recognition" [8].

## 2. DATA DESCRIPTION & PREPARATION

The dataset maps the first 1,024 bytes of firmware code for malicious, benign, and hacked Internet of Things (IoT) firmware from embedded software binaries such as Executable and Linkable Format (ELF) to thumbnail image format (Figure 1). Images are saved as csv files, in which "filename, label class… and the first 1024 bytes are mapped to grayscale" [8] by converting individual bytes to a decimal from 0 to 15 and then scaled. Initial probing showed great promise in metrics like F1 and accuracy scores. As IoT security evolves, one focus must be on reducing the number of false negatives due to the potential damage of missing an attack. Reducing false positives will aid in better identification of malware in general.

EDA quickly reveals the discrepancy between class populations [9]. The three classes represented are benignware, hackware, and malware.  There are a total of 38,8887 samples: 38,073 benignware examples, while hackware and malware have 103 and 711 examples, respectively (see Figure 2). The different types of code represented are all similar in structure, thus making immediate visual discrimination of class practically impossible (see Figure 3).

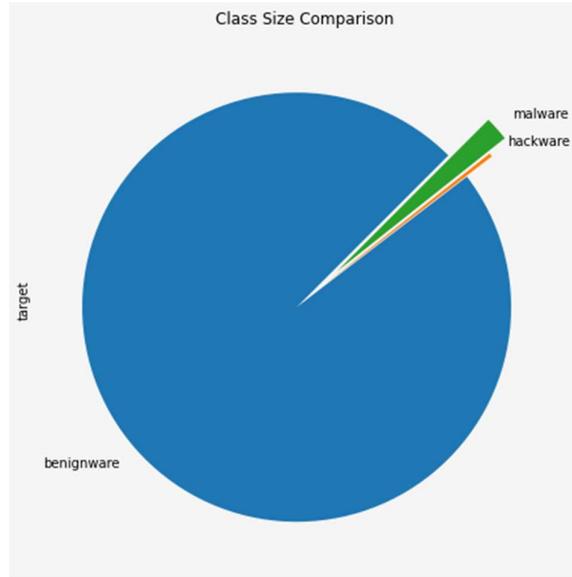

Figure 2. The csv data contains a large class population discrepancy

Such large differences are known to decrease prediction scores but can be dealt with in numerous ways including stratification, bootstrapping, over/under-sampling, and the Synthetic Minority Over-sampling Technique (SMOTE) which creates synthetic data that "fixes" the imbalance [10]. This method is used in the next section.

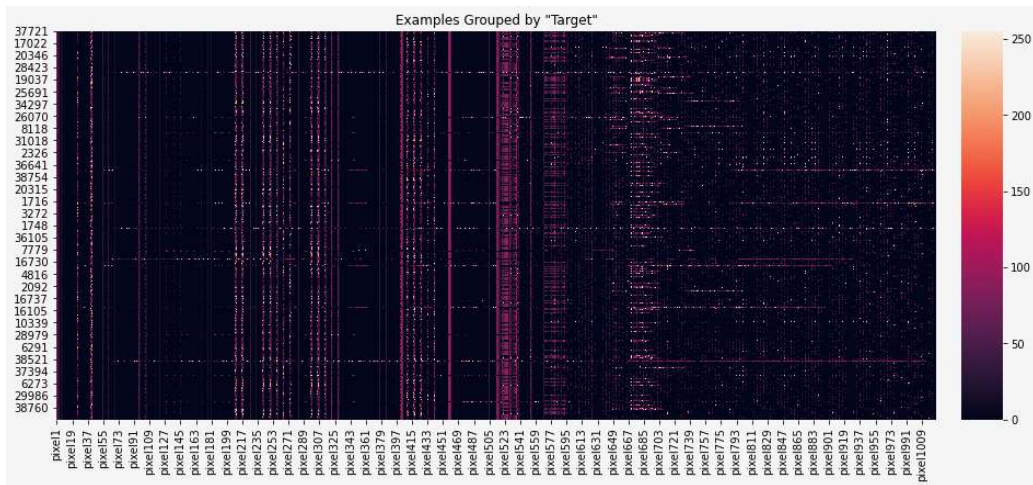

Figure 3. Heatmap visualization shows little differentiation between classes

## 3. MODEL COMPARISON RESULTS

### 3.1. PyCaret

A general comparison using the PyCaret module [11] with no hyperparameter tuning yields excellent results across 13 ML models (Table 1). PyCaret employs SMOTE to smooth the population imbalance, then trains with 3-fold cross-validation and pixel normalization, but no hyperparameters are tuned. Table 1 depicts accuracy and F1 scores of 100% for Gradient Boosting Classifier. Logistic Regression (LR) and Random Forest Classifier (RF) have F1 scores of 0.9999 and both show unity in Area Under the ROC Curve (AUC). With models performing comparably well, time becomes the most important metric when an attack occurs. In this case LR has a temporal cost that is an order of magnitude less than Gradient Boosting, and RF is five times less than even that, taking slightly over six seconds to complete. All show comparable performance with RF having the lowest recall score of the three.

Table 1. PyCaret Model Comparison Results.

| Model | Accuracy | AUC | F1 | Kappa | MCC | TT (Sec) |
|---|---|---|---|---|---|---|
| Gradient Boosting Classifier | 1.0000 | 1.0000 | 1.0000 | 0.9990 | 0.9990 | 318.7067 |
| Logistic Regression | 0.9999 | 1.0000 | 0.9999 | 0.9971 | 0.9971 | 31.9933 |
| Random Forest Classifier | 0.9999 | 1.0000 | 0.9999 | 0.9961 | 0.9961 | 6.1133 |
| Light Gradient Boosting Machine | 0.9999 | 1.0000 | 0.9999 | 0.9961 | 0.9961 | 18.7200 |
| Extra Trees Classifier | 0.9998 | 1.0000 | 0.9998 | 0.9951 | 0.9951 | 8.2167 |
| K Neighbors Classifier | 0.9994 | 0.9999 | 0.9994 | 0.9847 | 0.9848 | 262.4367 |
| Decision Tree Classifier | 0.9989 | 0.9910 | 0.9989 | 0.9722 | 0.9723 | 3.1100 |
| SVM – Linear Kernel | 0.9986 | 0.0000 | 0.9985 | 0.9609 | 0.9615 | 2.7600 |
| Ridge Classifier | 0.9907 | 0.0000 | 0.9918 | 0.8016 | 0.8178 | 2.0133 |
| Linear Discriminant Analysis | 0.9891 | 0.9907 | 0.9907 | 0.7746 | 0.7949 | 13.0400 |
| Quadratic Discriminant Analysis | 0.9852 | 0.4489 | 0.9797 | 0.3399 | 0.3898 | 10.5600 |
| Naïve Bayes | 0.8980 | 0.9444 | 0.9337 | 0.2490 | 0.3765 | 1.6967 |
| Ada Boost Classifier | 0.8397 | 0.8539 | 0.8936 | 0.1651 | 0.2243 | 24.4967 |

### 3.2. Convolutional Neural Network (CNN)

The CNN framework proved very effective with little architecting or hyper-parameter tuning. A simple model with only one convolutional layer using 16 filters and one fully connected (FC) hidden layer with 1024 nodes was able to achieve unity in all metrics after only five iterations. Overfitting is not seen in the test scores due to dropout (0.3) between and L1/L2 regularization (0.0001) on these backend FC layers.

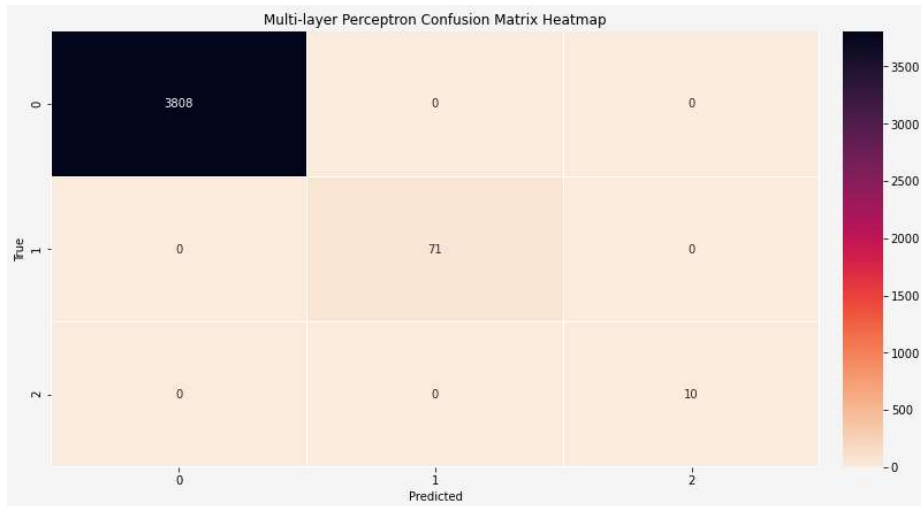

Figure 4. MLP perfects classifications while the support highlights population differences

### 3.3 Dense Neural Network & Multi-layer Perceptron

Both classifiers – from Keras version 2.4.0 and Scikit-Learn, respectively – performed well with minimal tuning and layering. Training scores reached unity within a few iterations, taking no more than 69.7 seconds for Keras' Dense Sequential network, and only 31.3 seconds for the Multi-layer Perceptron (MLP).

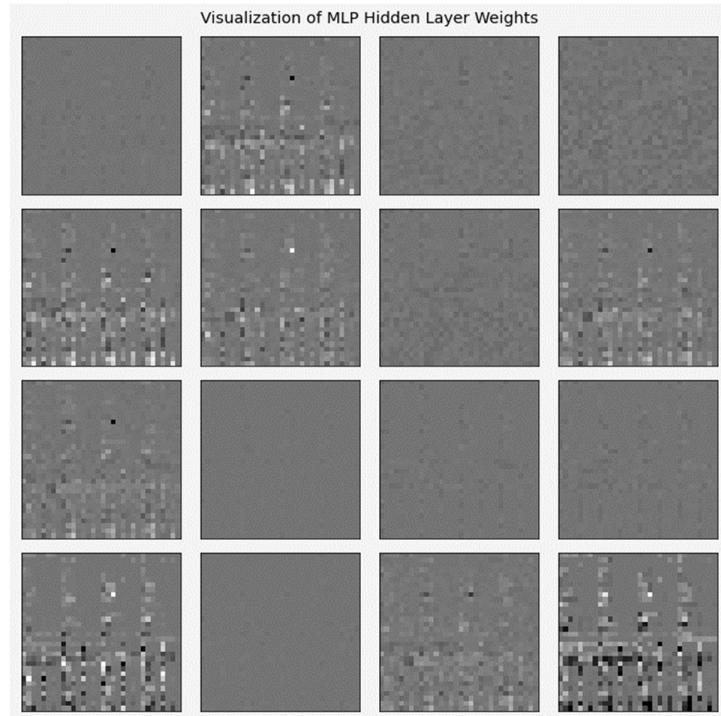

Figure 5. Hidden layer neurons showing feature detection

Predictions showed high fidelity in each: only one misclassification in the Dense model, and perfect unity on test data for the MLP (Table 2). These excellent results and low processing times make both models contenders for deployment, with MLP and RF having the least temporal cost.

### 3.4 Graph Convolutional Network (GCN)

A GCN was initially trained on the entirety of the dataset with a 70/30 train-test split. However, the significant class imbalance resulted in the GCN classifying all samples as the most prevalent class. To resolve this, we used random sampling and under-sampling of the classes with larger sample sizes to ensure a balanced class mixture. This was then divided into the training and test rations mentioned above.. The resulting adjacency graphs for each of the classes were well-differentiated, and the network was able to reach an accuracy of 92% after 100 epochs of training, which was completed in approximately 30 seconds.

Table 2. CNN, FC DNN, MLP & GCN Results

| Model | Accuracy | F1 | MCC | TT (Sec) |
|---|---|---|---|---|
| Convolutional Neural Network | 1.0 | 1.0 | 1.0 | 420.834 |
| Dense Sequential Network | 1.0 | 0.99 | 0.9958 | 69.7176 |
| Multi-layer Perceptron | 1.0 | 1.0 | 1.0 | 31.2957 |
| Graph Convolutional Network | 0.92 | 0.92 | – | ~30 |

## 4. MODEL OPTIMIZATIONS

In this section the top three models from the comparison shown in Table 1 are optimized via hyperparameter grid search and 3-fold cross-validation. Normalization is performed by maximum pixel value division yielding pixel values $v \in [0 – 255]$. Balanced accuracy and the macro average of F1 score are used as success metrics. The Matthews Correlation Coefficient (MCC) is used to assess the quality of the predictions and hence, the viability of the results.

### 4.1 Gradient Boosting Classifier

This model continues to produce excellent predictions: all scores training reach unity including MCC. Overfitting is not seen when the model is evaluated on test data, scoring 0.987065 in balanced accuracy and 0.991277 F1 score. Gradient boosting is known to be time consuming, and this is seen here with a mean fit time of 569.51 seconds and mean score time of 0.69976 seconds.

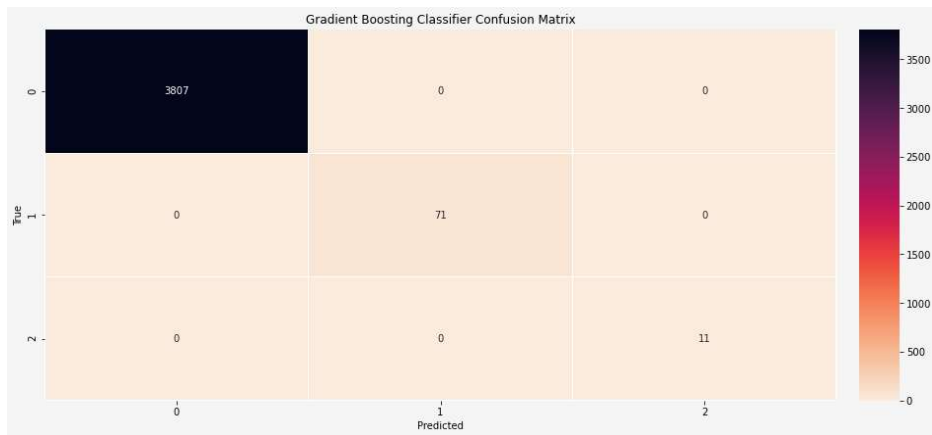

Figure 6. Final results heatmap

### 4.2 Logistic Regression Classifier

This model improves as the hyper-parameter C – the inverse of the regularization parameter λ – increases. Thus, less regularization on LR is beneficial for this task. The L2 penalty is applied, and the best solver proved to be lbfgs. During the final test one malware example was misclassified as benignware, leading to recall and F1 scores of 0.99. The MCC reached 0.99377 while balanced accuracy achieved 0.99531. The final macro average for F1 is 0.99759.

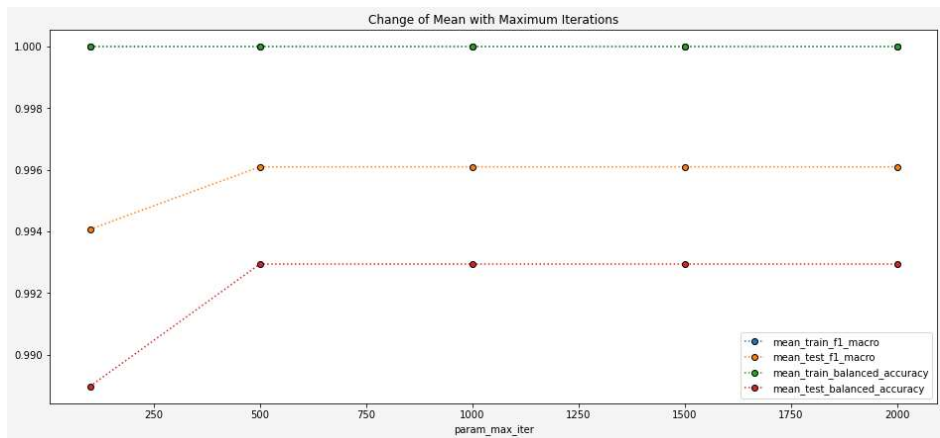

Figure 7. Scores plateau with increasing iterations

## 4.3 Random Forest Classifier

Once again this model achieves excellent scores in all category and displays no overfitting with all training and validation metrics achieving 100%. Using 1000 estimators and entropy criterion, it proves to be the fastest to train, taking an average of 9.35 seconds fit time on the test data. Final test balanced accuracy and F1 macro average are 0.99179 and 0.99580, respectively.

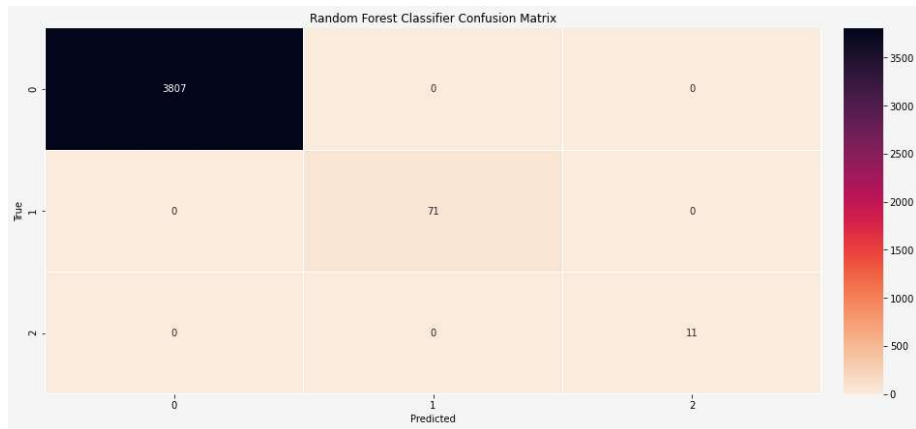

Figure 8. RF shows no confusion on validation data

## 5. CONCLUSIONS

The methods tested produce excellent results for identifying each type of malicious code. Little can be done to improve their scores, making detailed optimization practically unnecessary. Grid search improved the random forest approach, making the RF and MLP classifiers the most viable candidates because of their low training times. The Matthews Correlation Coefficient (MCC), or phi coefficient, bolsters this assessment with scores at or near unity, indicating that the predictions are high quality and using them is much better than simply guessing. In such cases where each algorithm performs well, time can become the deciding metric. An algorithm that can produce such outstanding results the fastest makes a more suitable candidate for deployment and defence of critical IoT infrastructure.

Such high scores in valuable metrics like accuracy, precision, and recall bode well for defence of future Internet of Things devices and networks. However, as malware adapts to these new image classification techniques there will arise a need for even more sophisticated binary file transformation and malicious code detection methods.

## ACKNOWLEDGEMENTS

The authors would like to thank the PeopleTec, Inc. Technical Fellows program for encouraging and assisting this research.

## Authors

Erik Larsen, M.S. is a senior data scientist with research experience in quantum physics and deep learning. He completed both M.S. and B.S. in Physics at the University of North Texas, and a B.S. in Professional Aeronautics from Embry-Riddle Aeronautical University while serving as an aviator in the U.S. Army.

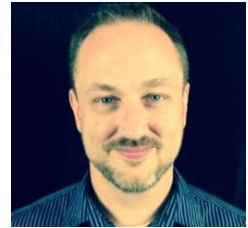

Korey MacVittie, M.S. is a data scientist specializing in machine learning. Prior research includes identifying undervalued players in sports drafting. He completed his M.S. at Southern Methodist University, and a B.S. in Computer Science, and a B.S. in Philosophy from the University of Wisconsin Green Bay.

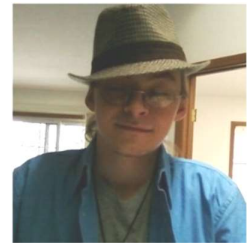

John R. Lilly III is an applied machine learning research scientist specializing in quantitative analysis, algorithm development, and implementation for embedded systems. Prior research includes cryptographic assets for multiple enterprise blockchain protocols. He has been trained in statistics by Cornell University. Formerly infantry, he currently serves as an intelligence advisor for the U.S. Army's Security Force Assistance Brigade.

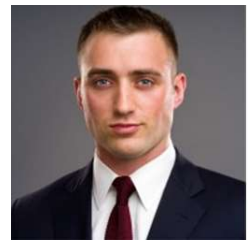